\documentclass[letterpaper, 10 pt, conference]{ieeeconf}

\IEEEoverridecommandlockouts   % Needed for \thanks
\usepackage[utf8]{inputenc}
\usepackage[T1]{fontenc}
\usepackage{microtype}
\usepackage{graphicx}
\usepackage{subcaption}
\usepackage{booktabs}
\usepackage{url}
\let\labelindent\relax   % ieeeconf already defines \labelindent; avoid clash with enumitem
\usepackage{enumitem}
\usepackage{amsmath,amssymb,amsfonts}
\usepackage{amsthm}
\usepackage{mathtools}
\usepackage{nicefrac}
\usepackage{bm}
\usepackage{multirow}
\usepackage{colortbl}
\usepackage{hhline}
\usepackage{caption}
\usepackage{algorithm}
\usepackage{algpseudocode}
\usepackage{array}
\usepackage{makecell}
\usepackage{wrapfig}
\usepackage{fancyhdr}
\usepackage{pifont}
\usepackage{xspace}
\usepackage{titletoc}
\usepackage{tcolorbox}
\usepackage{tablefootnote}
\usepackage{fontawesome}
\usepackage{xcolor}

\makeatletter\let\NAT@parse\undefined\makeatother
\usepackage[numbers,sort&compress]{natbib}

\usepackage[capitalize,noabbrev]{cleveref}

\usepackage[disable,textsize=tiny]{todonotes}
\definecolor{lightred}{rgb}{0.988, 0.294, 0.0823}
\definecolor{lightblue}{rgb}{0.11, 0.541, 0.752}
\definecolor{lightgreen}{rgb}{0.1, 0.6, 0.1}

\theoremstyle{plain}

\theoremstyle{definition}

\theoremstyle{remark}

\title{\LARGE \bf
AWARE: Action-Entropy-Weighted and Attention-Guided Retrieval of
Observation Evidence for Vision-Language-Action Models
}

\author{Jiabing Yang$^{1,2,5}$, Yixiang Chen$^{1,2}$, Yuan Xu$^{1,2}$, Peiyan Li$^{1,2}$, Zichen Wen$^{3}$, Bowen Fang$^{1,2}$, \\
Chenhang Cui$^{4}$, Tao Yu$^{1,2}$, Qisen Ma$^{1,2}$, Kai Wang$^{1,2}$, Ziheng He$^{1,2}$, Wen Ye$^{1,2}$, \\
Xiangnan Wu$^{1,2}$, Yingda Li$^{1,2}$, Yan Huang$^{1,2,5,\dag}$, Liang Wang$^{1,2}$%
\thanks{$^{\dag}$Corresponding author.}%
\thanks{$^{1}$School of Artificial Intelligence, University of Chinese Academy of Sciences.}%
\thanks{$^{2}$New Laboratory of Pattern Recognition (NLPR), Institute of Automation, Chinese Academy of Sciences.
{\tt\small yangjiabing2025@ia.ac.cn, yhuang@nlpr.ia.ac.cn}}%
\thanks{$^{3}$Shanghai Jiao Tong University.}%
\thanks{$^{4}$National University of Singapore.}%
\thanks{$^{5}$FiveAges.}%
}

\newcommand{\method}{\textsc{AWARE}\xspace}
\newcommand{\methodfull}{\textbf{A}ction-Entropy-\textbf{W}eighted and \textbf{A}ttention-Guided \textbf{R}etrieval of Observation \textbf{E}vidence\xspace}
\ifdefined\VARIANTI\variantitrue\fi

\begin{document}
\maketitle
\begin{abstract}
Vision-Language-Action (VLA) models combine multimodal perception with robotic control, demonstrating remarkable potential for generalizable manipulation. Yet, enhancing their execution typically relies on auxiliary sensory inputs (e.g., depth, 3D point clouds) or costly fine-tuning. Inspired by the insight that Feed-Forward Networks (FFNs) act as key-value memory, we propose \methodfull~(\method), an effective, training-free, and plug-and-play module for VLAs. Specifically, \method tracks layer-wise uncertainty via an element-wise \textbf{\emph{Action Entropy}} metric, and retrieves observation evidence into the element\textquotesingle s FFN through associative memory lookup. Crucially, the retrieved features are weighted by instruction-guided cross-modal attention to isolate task-relevant observation cues, while injection strength is adaptively scaled by the element\textquotesingle s entropy and relative network depth. This selectively channels observation evidence to high-entropy action dimensions and deep layers where grounding decays, while leaving converged representations intact. Across LIBERO, SIMPLER, CALVIN, and real-world robot manipulation, \method consistently improves diverse VLA architectures with minimal computational overhead, achieving the largest gains on long-horizon tasks without architectural changes or additional training. Project page: \url{https://jiabingyang01.github.io/AWARE/}.
\end{abstract}

\section{Introduction}
Recent advancements in Vision-Language Models (VLMs)~\cite{Paligemma, Qwen2.5-vl, Prismatic} have catalyzed the development of Vision-Language-Action (VLA) models~\cite{OpenVLA, pi0, OpenVLA-OFT}, which map visual observations and language instructions into low-level robot actions. Trained on web-scale pretraining corpora and demonstration data, modern VLAs generalize across diverse manipulation settings. Nonetheless, further boosting policy performance typically incurs prohibitive costs, constrained by the scarcity of physical robot data and non-negligible computational overhead required for fine-tuning.

To enhance manipulation precision without massive data collection, prevailing approaches frequently augment visual inputs with specialized priors, such as visual trace prompting~\cite{Tracevla}, 3D spatial encodings~\cite{SpatialVLA}, or depth and point-cloud modalities~\cite{3D-CAVLA, Evo-0, AimBot}. Although effective, these interventions necessitate customized sensor configurations, auxiliary encoders, and extensive task-specific fine-tuning. This motivates an appealing yet challenging question: \textit{can we boost existing VLA policies entirely at inference time, in a training-free manner and without supplementary observation cues, external sensors, or auxiliary network modules?}

\begin{figure}[t]
  \centering
  \includegraphics[width=\linewidth]{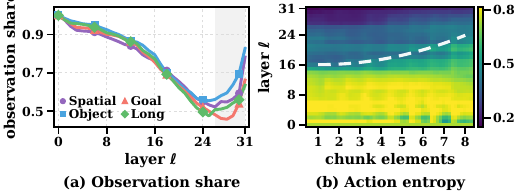}
  \caption{\textbf{Empirical motivation for \method} (OpenVLA-OFT on LIBERO). \textbf{(a)} Observation subspace footprint diminishes across network depth, halving by layer 26 across all suites. \textbf{(b)} Action Entropy across all 56 action chunk elements: near steps settle early in middle layers while distant steps retain elevated predictive entropy until deep layers, forming a clear decision wavefront (white dashed isocontour) that motivates element-wise observation routing.}
  \label{fig:mechanism}
\end{figure}

To address this, we investigate the internal representation dynamics of VLA models during forward inference. Two phenomena stand out. First, as tokens traverse the transformer stack, the proportion of the action representation explained by observation tokens diminishes monotonically, losing half of its subspace footprint by layer 26 across all task suites (Fig.~\ref{fig:mechanism}a). Second, predictive uncertainty resolves asynchronously across action dimensions: early trajectory steps converge early in the layer stack, whereas as shown in Fig.~\ref{fig:mechanism}b, distant future steps maintain elevated predictive entropy until deep layers, exhibiting a persistent uncertainty gap of $0.25$ across the prediction horizon. A natural remedy suggests itself: re-inject observation evidence where predictive uncertainty remains high. Because Feed-Forward Networks (FFNs) act as key-value associative memories~\cite{geva2021transformer, jie2024memory}, the model\textquotesingle s native FFN layers can dynamically retrieve observation cues and reinforce action representations in the middle of the stack.

Building on these observations, we propose \methodfull~(\method), a lightweight, training-free, and plug-and-play module that lets each action element autonomously regulate its own observation retrieval. \method computes the layer-wise \textbf{\emph{Action Entropy}} for every individual action dimension, in place of a coarse aggregate scalar. It queries observation tokens using each element\textquotesingle s hidden state and blends the retrieved features into the FFN output, modulated by the element\textquotesingle s own entropy and normalized network depth. This continuous formulation channels observation evidence specifically to high-entropy action dimensions and deep layers where perception has faded, while preserving converged representations intact, with the retrieval conditioned by the instruction\textquotesingle s cross-modal attention. In summary, our key contributions are:

\begin{itemize}[leftmargin=10pt, topsep=0pt, itemsep=1pt, partopsep=1pt, parsep=1pt]
\item We formulate an element-wise \textbf{\emph{Action Entropy}} metric to track layer-wise predictive uncertainty, revealing pronounced divergence across temporal horizons.
\item We present \textbf{\method}, a training-free framework that uses native FFN memory to re-inject observation evidence into individual action elements proportionally to their uncertainty and depth, guided by instruction-relevant visual attention.
\item We validate \method across standard simulation benchmarks (LIBERO, SIMPLER, CALVIN) and real-world robots. The gains hold across diverse policy architectures, and ablations trace them to element-wise allocation rather than arbitrary perturbation.
\end{itemize}

\section{Related Work}
\label{sec:related_work}

\noindent \textbf{Vision-Language-Action Models.} VLA models fine-tune pretrained VLMs on robot data to map observations and language onto low-level control~\cite{Rt-1,Rt-2,OpenVLA,Cogact,pi0,BridgeVLA++}, a recipe that scales with the cross-embodiment corpora behind it~\cite{OXE,Octo}. They differ in how the action is represented, and that is what decides where an inference-time rule can attach: OpenVLA~\cite{OpenVLA} decodes discrete action tokens through its language head, OpenVLA-OFT~\cite{OpenVLA-OFT} decodes a whole chunk of placeholder tokens in parallel, CogACT~\cite{Cogact} conditions a diffusion expert on a single cognition token, and $\pi_0$~\cite{pi0} attaches a flow-matching expert to a PaliGemma prefix~\cite{Paligemma}. Another line pretrains action representations from video data~\cite{LAPA,Univla,EC-Flow,BridgeV2W}. \method{} is defined on the positions each family reads its action from, so one rule covers all of them.

\noindent\textbf{Uncertainty in Robot Learning.} Output entropy has guided dynamic decoding and token pruning in language models~\cite{fang2024uncertainty}, and internal confidence has guided test-time visual retrieval to mitigate perceptual fading~\cite{zou2024look, yang2025ikod}. In robotic control, \citet{valle2025evaluating} read token entropy as a VLA confidence score, and \citet{karli2025ask} use it to trigger human intervention. These signals gate a whole token, a whole layer, or a whole observation. A chunked VLA emits every step of a trajectory in one pass and its elements are at different stages of being decided within that pass, so the useful signal is not one number per layer but one per element, which is what \method{} acts on.

\noindent\textbf{Perceptual Enhancement for Manipulation.} A common remedy for weak visual grounding is to add something to the input, such as visual traces~\cite{Tracevla}, point clouds and depth~\cite{3D-CAVLA}, three-dimensional priors from a geometry model~\cite{Evo-0}, or overlaid aiming cues~\cite{AimBot}. Others keep the input and change what the policy is trained to attend to, through gaze reconstruction~\cite{reconvla} or attention specialised to task-relevant factors~\cite{GuidedVLA}. Each buys accuracy with a sensor, an auxiliary encoder or task-specific fine-tuning. \method{} adds nothing: it re-reads the observation the model has already encoded, through the model's own FFN, at inference time.

\section{Methodology}
\label{sec:method}

\begin{figure*}[t]
    \centering
    \includegraphics[width=\linewidth]{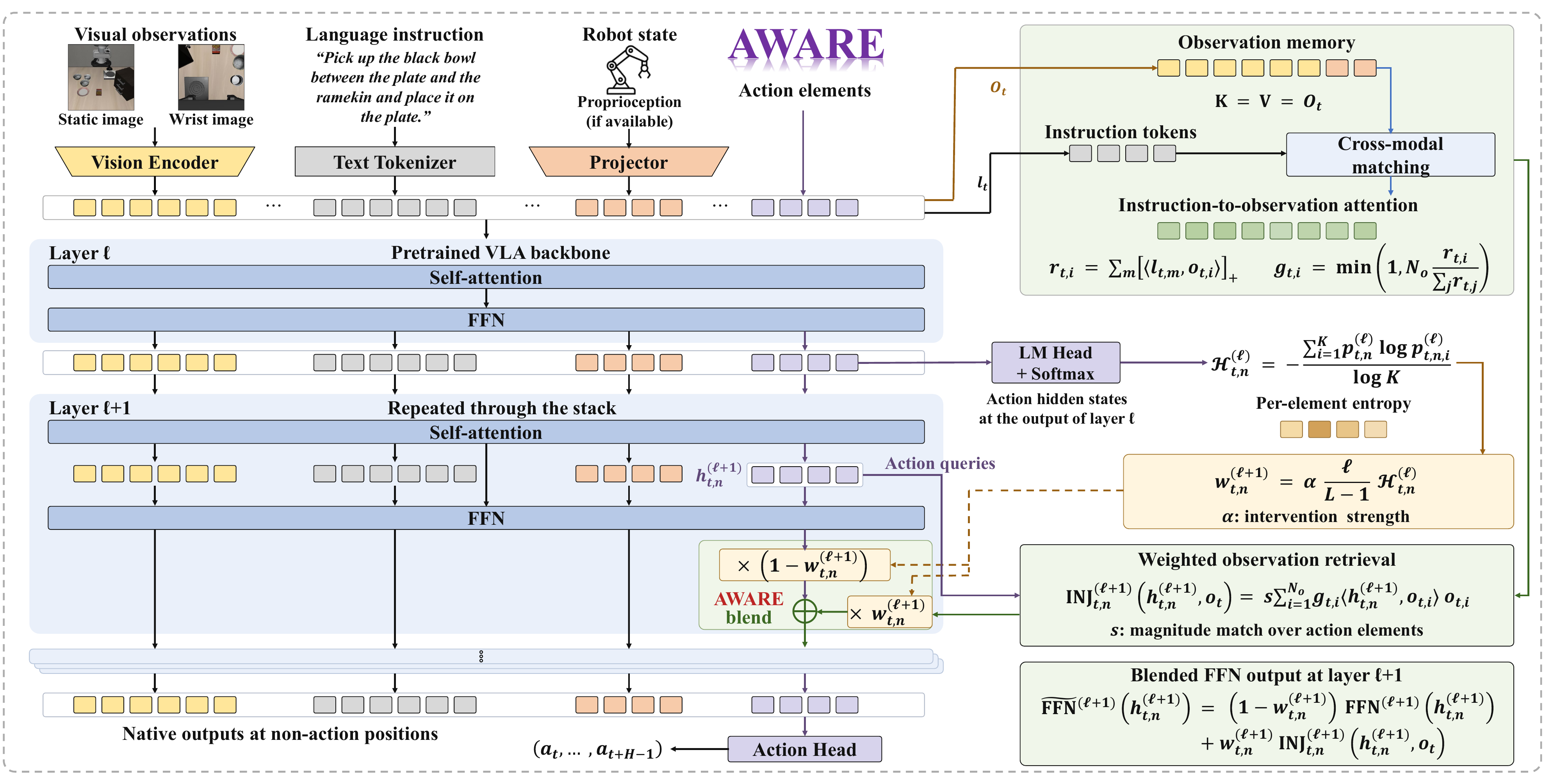}
    \caption{\textbf{Overall framework of \method.} At every layer the language head reads each element of the predicted action representations and gives its Action Entropy. The observation features form a key-value memory that the element queries through its FFN input, and the retrieved features enter its FFN output at the next layer with weight $w^{(\ell+1)}_{t,n}=\alpha\,\tfrac{\ell}{L-1}\,\mathcal{H}^{(\ell)}_{t,n}$. Nothing is switched on or off, and the memory is read at every layer.}
    \label{fig:aware_pipeline}
\end{figure*}

\subsection{Feed-Forward Networks as Associative Memory}
\label{sec:preliminary}
Following the continuous associative memory formalism of \citet{geva2021transformer}, a transformer feed-forward network (FFN) with weights $\boldsymbol{W}_1 \in \mathbb{R}^{d\times D}$, $\boldsymbol{W}_2 \in \mathbb{R}^{D\times d}$ and activation $\phi$ computes:
\begin{equation}
\operatorname{FFN}(\boldsymbol{h}) \;=\; \phi(\boldsymbol{h}\,\boldsymbol{W}_1)\,\boldsymbol{W}_2 \;=\; \sum_{i=1}^{D}\phi(\langle \boldsymbol{h}, \boldsymbol{k}_i\rangle)\,\boldsymbol{v}_i,
\label{eq:1}
\end{equation}
where the parameter columns of $\boldsymbol{W}_1$ and $\boldsymbol{W}_2$ constitute $D$ key-value pairs $\{(\boldsymbol{k}_i, \boldsymbol{v}_i)\}_{i=1}^{D}$ ($\boldsymbol{k}_i, \boldsymbol{v}_i \in \mathbb{R}^d$). Under this key-value view, the FFN executes an unnormalized soft associative retrieval: the hidden state $\boldsymbol{h}$ acts as a query against stored keys $\boldsymbol{k}_i$, aggregating matching value vectors $\boldsymbol{v}_i$ weighted by inner-product similarity. Building upon this formulation, recent vision-language studies demonstrate that transformer memory spaces can be dynamically conditioned at inference time with external multimodal tokens to mitigate perceptual fading and attention degradation~\cite{jie2024memory, zou2024look, yang2025ikod}. In robotic manipulation, where multi-step action execution is vulnerable to cumulative observation fading (Fig.~\ref{fig:mechanism}a), we exploit this native associative structure to selectively retrieve instruction-grounded observation evidence directly into intermediate action representations, without altering pretrained parameters.

\subsection{Problem Formulation}
Vision-Language-Action (VLA) models are designed to jointly process observations and language instructions for the purpose of generating appropriate actions for robots. Formally, given the observation $\boldsymbol{o}_t$ at time $t$ and language instruction $\boldsymbol{l}$, a model $\boldsymbol{\pi}$ predicts a temporal action sequence $(\boldsymbol{a}_t,\boldsymbol{a}_{t+1},\ldots,\boldsymbol{a}_{t+H-1})$ (i.e., action chunk size $H$) for task execution:
\begin{equation}
    \boldsymbol{\pi}: (\boldsymbol{o}_t, \boldsymbol{l}) \rightarrow (\boldsymbol{a}_t,\boldsymbol{a}_{t+1},\ldots,\boldsymbol{a}_{t+H-1}). \label{eq:4}
\end{equation}
The observation is the visual input, concatenated with the proprioceptive state if available~\cite{pi0, OpenVLA-OFT}, and each $\boldsymbol{a}_t$ is a 7-DoF end-effector command: three translations, three rotations and one gripper state. A chunked policy such as OpenVLA-OFT represents the $N_a = H D_a$ action elements ($H{=}8$, $D_a{=}7$, indexed by $n \in \{1, \ldots, N_a\}$) as $N_a$ placeholder tokens that are decoded in parallel in one bidirectional forward pass.

\subsection{Uncertainty-Aware Observation Retrieval}
\label{sec:aware}

\noindent\textbf{Element-Wise Action Entropy Formulation.} Building on prior inference-time interventions that exploit layer-wise predictive uncertainty in language models~\cite{chuang2024dola, fang2024uncertainty}, we tailor an element-wise \textbf{\emph{Action Entropy}} metric $\mathcal{H}_{t,n}^{(\ell)}$ for chunked VLAs, quantifying decision uncertainty via the entropy of action-related output distributions across the layer hierarchy. The quantity is the entropy of the vocabulary distribution read at the $N_a$ positions the action pathway consumes: the action tokens themselves where actions come straight from hidden states~\cite{OpenVLA-OFT}, the condition token where a second system decodes them~\cite{Cogact}, or the prefix tokens where a flow-matching expert attends to them. We define the layer-wise action entropy of the $n$-th action element ($n \in \{1, \ldots, N_a\}$) at time step $t$ as
\begin{equation}
    \mathcal{H}^{(\ell)}_{t,n} = -\,\frac{\sum_{i=1}^{K} p^{(\ell)}_{t,n,i}\,\log p^{(\ell)}_{t,n,i}}{\log K}.
\label{eq:6}
\end{equation}
Here $\bm{p}^{(\ell)}_{t,n}=\{p^{(\ell)}_{t,n,i}\}_{i=1}^K$ is the categorical distribution over the candidate tokens for the $n$-th action or condition token, obtained by projecting the element's hidden state at the output of layer $\ell$ through the language modeling head and normalizing with softmax, following the ``Logit Lens'' formulation~\cite{logic_lens, tuned_lens}. For discrete actions, we set $K=256$ to match the number of action bins; for continuous actions, we likewise fix $K=256$ for definitional consistency. The normalization ensures $\mathcal{H}^{(\ell)}_{t,n}\in[0,1]$. \method retains all $N_a$ entropy values, enabling fine-grained, element-specific modulation across the action dimensions.

\noindent\textbf{Interpretation of Action Entropy.} In Transformer architectures, representations evolve via iterative refinement along the residual stream. Extensive work in mechanistic interpretability~\cite{logic_lens, tuned_lens} and contrastive decoding~\cite{chuang2024dola} establishes that projecting intermediate hidden states through the frozen language head reliably elicits latent categorical predictions without auxiliary training, correlating strongly with policy execution uncertainty in robotic manipulation~\cite{valle2025evaluating, karli2025ask}.
A potential concern is representation drift: intermediate hidden states are not explicitly trained against the final unembedding head, which could introduce an unaligned coordinate basis. Crucially, as established in~\cite{tuned_lens}, any such affine shift is \emph{spatially invariant across tokens at the same layer $\ell$}. Such layer-wide drift manifests solely as an additive scalar offset to the layer's baseline entropy, which is absorbed by the depth schedule $\ell/(L{-}1)$ and scaling factor $\alpha$. \method never relies on absolute cross-layer entropy calibration; it operates strictly on the \emph{relative entropy contrast among action elements within the same layer $\ell$} (the decision wavefront in Fig.~\ref{fig:mechanism}b). Because all $N_a$ action elements share the identical coordinate geometry at layer $\ell$, their intra-layer relative entropy differences isolate genuine decision uncertainty, a mechanism directly corroborated by our diagnostic experiments in Table~\ref{tab:controls}.

\noindent\textbf{Cross-Backbone Token Mapping.} The core formulation (Eqs.~\ref{eq:6}--\ref{eq:8}) remains identical across all backbones. The only model-dependent adaptation is the token positions where uncertainty is evaluated, aligning with how each architecture structures its action pathway. For chunked policies with bidirectional attention such as OpenVLA-OFT, entropy is evaluated across all $N_a$ action placeholder tokens simultaneously ($N_a{=}56$). For causal autoregressive models (OpenVLA, SpatialVLA), it is computed at each newly decoded action token ($N_a{=}1$). In dual-system architectures like CogACT, it is evaluated at the cognition token that conditions the action decoder. Finally, for flow-matching policies whose continuous expert lacks a vocabulary head ($\pi_0$, $\pi_{0.5}$), the entropy is evaluated at the instruction positions of the prefix that condition the expert. We restrict this evaluation to the instruction tokens rather than the full prefix because visual patch tokens already constitute the observation memory $\boldsymbol{o}_t$, and the language head yields meaningful probability distributions only over text tokens.

\noindent\textbf{Element-Wise FFN Retrieval.} To provide timely perceptual grounding for action dimensions that remain unresolved during execution, \method retrieves observation evidence into the subsequent FFN layer, allowing each action element to autonomously regulate its own retrieval. We treat the encoded observation features $\boldsymbol{o}_t=(\boldsymbol{o}_{t,1}, \ldots, \boldsymbol{o}_{t,N_o})$ as a key-value memory and use the hidden state $\boldsymbol{h}_{t,n}^{(\ell+1)}$ entering the FFN of action element $n$ at layer $\ell{+}1$ as the query:
\begin{equation}
    \textsc{inj}_{t,n}^{(\ell+1)}\bigl(\boldsymbol{h}_{t,n}^{(\ell+1)}, \boldsymbol{o}_t\bigr)
    = s\sum_{i=1}^{N_o} g_{t,i}\,\langle \boldsymbol{h}_{t,n}^{(\ell+1)}, \boldsymbol{o}_{t,i} \rangle \, \boldsymbol{o}_{t,i}.
    \label{eq:9}
\end{equation}
Here $g_{t,i}$ is the instruction-to-observation attention defined below, and the retrieval operates as a linear associative memory lookup following Eq.~\ref{eq:1} with identity activation. In this formulation, observation features act simultaneously as keys and values, and the query $\boldsymbol{h}_{t,n}^{(\ell+1)}$ retrieves a linear combination of observation tokens weighted by semantic alignment and instruction relevance. In practice, Eq.~\ref{eq:9} is vectorized across tokens as $s\,\boldsymbol{O}_t^{\!\top}\operatorname{diag}(\boldsymbol{g}_t)\,\boldsymbol{O}_t\,\boldsymbol{h}_{t,n}^{(\ell+1)}$ where $\boldsymbol{O}_t\!\in\!\mathbb{R}^{N_o\times d}$ stacks the observation tokens, requiring only two standard matrix multiplications without iterative looping or matrix inversion. The scaling factor $s$ matches the mean activation magnitude of the native FFN to that of the retrieved features over the action elements, ensuring well-conditioned feature blending. The retrieved evidence is then integrated into the FFN output of element $n$ according to its own uncertainty.

\noindent\textbf{Instruction-Guided Observation Conditioning.}
Without task conditioning, an unconstrained associative query $\langle \boldsymbol{h}_{t,n}^{(\ell+1)}, \boldsymbol{o}_{t,i}\rangle$ can disperse broadly over irrelevant visual background or distractors with high token norms. Because robot manipulation is fundamentally governed by the language instruction specifying target objects and spatial relations, we modulate retrieval weights via cross-modal semantic alignment between instruction tokens and observation patches:
\begin{equation}
g_{t,i} \;=\; \min\!\Bigl(1,\; N_o\,\frac{r_{t,i}}{\sum_{j=1}^{N_o} r_{t,j}}\Bigr),\quad
r_{t,i}=\sum_{m}\bigl[\langle \boldsymbol{l}_{t,m}, \boldsymbol{o}_{t,i}\rangle\bigr]_{+}.
\label{eq:g}
\end{equation}
Here, $r_{t,i}$ aggregates non-negative semantic alignment across instruction tokens. Comparing against the uniform baseline $1/N_o$ yields a normalized importance score: observation tokens receiving at or above uniform attention retain full retrieval weight ($g_{t,i}{=}1$), while task-irrelevant regions are smoothly attenuated. Computing $\boldsymbol{g}_t$ requires only a single matrix multiplication per forward pass with zero learnable parameters, focusing associative retrieval on instruction-grounded visual entities.

\noindent\textbf{Continuous Adaptive Blending.} Element $n$ at layer $\ell{+}1$ carries the weight
\begin{equation}
w^{(\ell+1)}_{t,n} \;=\; \alpha\,\frac{\ell}{L-1}\,\mathcal{H}^{(\ell)}_{t,n},
\label{eq:w}
\end{equation}
read from the entropy of the layer below and applied to the FFN of this one, scaled by the intervention strength $\alpha > 0$. The normalized depth ratio $\ell/(L{-}1)$ progressively scales the intervention, mirroring the empirical decay of observation features across depth (Fig.~\ref{fig:mechanism}a), while the entropy factor $\mathcal{H}^{(\ell)}_{t,n}$ concentrates retrieval specifically on action elements with elevated predictive entropy. In practice, because distinct manipulation domains exhibit varying horizon lengths, visual clutter, and precision requirements, policy backbones naturally manifest different baseline action entropy distributions. To achieve optimal performance in practical deployment, $\alpha$ can be configured to match task characteristics, while remaining robust across a broad interior range (Section~\ref{sec:ablation}). The modified FFN output is then computed as:
\begin{equation}
\begin{aligned}
\widetilde{\operatorname{FFN}}{}^{(\ell+1)}\!\bigl(\boldsymbol{h}_{t,n}^{(\ell+1)}\bigr)
&= \bigl(1-w^{(\ell+1)}_{t,n}\bigr)\operatorname{FFN}^{(\ell+1)}\!\bigl(\boldsymbol{h}_{t,n}^{(\ell+1)}\bigr) \\
&\;+ w^{(\ell+1)}_{t,n}\, \textsc{inj}_{t,n}^{(\ell+1)}\!\bigl(\boldsymbol{h}_{t,n}^{(\ell+1)}, \boldsymbol{o}_{t}\bigr),
\end{aligned}
\label{eq:8}
\end{equation}
and $\widetilde{\operatorname{FFN}}{}^{(\ell+1)}\!\bigl(\boldsymbol{h}_{t,n}^{(\ell+1)}\bigr)$ replaces that element's native FFN output in the residual stream. Non-action tokens remain untouched. Because $w^{(\ell+1)}_{t,n}$ adapts continuously, settled action dimensions experience negligible modification, while unresolved dimensions in deeper layers receive targeted observation evidence.

\begin{algorithm}[!t]
\caption{\methodfull{} (\method).}
\label{algo:aware}
\begin{algorithmic}[1]
\Require VLA model $\boldsymbol{\pi}$ with $L$ layers; observation $\boldsymbol{o}_t{=}(\boldsymbol{o}^v_t,\boldsymbol{o}^p_t)$; instruction $\boldsymbol{l}$; strength $\alpha$.
\Ensure Action chunk $\boldsymbol{A}_t{=}(\boldsymbol{a}_t,\ldots,\boldsymbol{a}_{t+H-1})$.
\State Form the observation memory $\boldsymbol{O}_t$ from the encoded observation tokens. \hfill\Comment{Eq.~\ref{eq:9}}
\State Weight its rows by the instruction's own read $g_{t,i}$, once per forward pass. \hfill\Comment{Eq.~\ref{eq:g}}
\For{$\ell = 0$ \textbf{to} $L-2$}
    \State Apply the LM head to the $N_a$ action elements at the output of layer $\ell$ to obtain $\boldsymbol{p}^{(\ell)}_{t,n}$.
    \State Compute Action Entropy $\mathcal{H}^{(\ell)}_{t,n}$ for every element $n$. \hfill\Comment{Eq.~\ref{eq:6}}
    \State At layer $\ell{+}1$, retrieve $\textsc{inj}_{t,n}^{(\ell+1)}$ for every action element from $\boldsymbol{O}_t$. \hfill\Comment{Eq.~\ref{eq:9}}
    \State Blend with weight $w^{(\ell+1)}_{t,n}=\alpha\,\tfrac{\ell}{L-1}\,\mathcal{H}^{(\ell)}_{t,n}$; leave every other token unchanged. \hfill\Comment{Eqs.~\ref{eq:w},~\ref{eq:8}}
\EndFor
\State Decode $\boldsymbol{A}_t$ from the final hidden state through the policy head.
\end{algorithmic}
\end{algorithm}

\section{Experiments}

\begin{table*}[!t]
  % \small
  \centering
  \caption{Performance comparison on the LIBERO benchmark. All baselines are evaluated under identical checkpoints and evaluation pipelines. Bold denotes the better of a pair. Per-suite intervention strengths $\alpha$ for Spatial, Object, Goal, and Long are: OpenVLA-OFT (0.04, 0.05, 0.10, 0.03), $\pi_0$ (0.03, 0.04, 0.05, 0.03), $\pi_{0.5}$ (0.01, 0.08, 0.01, 0.07), and OpenVLA (0.03, 0.02, 0.01, 0.01).}
  % \resizebox{0.8\linewidth}{!}{
    \begin{tabular}{lccccc}
      \toprule
      \textbf{Method} & \textbf{Spatial} & \textbf{Object} & \textbf{Goal} & \textbf{Long} & \textbf{Average} \\
      \midrule
      % Reference baselines commented out for pure Base vs. w/ AWARE comparison
      % TraceVLA~\cite{Tracevla}~\textcolor{gray}{(\textit{ICLR'25})} & 84.6 & 85.2 & 75.1 & 54.1 & 74.8 \\
      % SpatialVLA~\cite{SpatialVLA}~\textcolor{gray}{(\textit{RSS'25})} & 88.2 & 89.9 & 78.6 & 55.5 & 78.1 \\
      % UniVLA~\cite{Univla}~\textcolor{gray}{(\textit{RSS'25})} & 96.5 & 96.8 & 95.6 & 92.0 & 95.2 \\
      % CogVLA~\cite{cogvla}~\textcolor{gray}{(\textit{NeurIPS'25})} & 98.6 & 98.8 & 96.6 & 95.4 & 97.4 \\
      % 3D-CAVLA~\cite{3D-CAVLA}~\textcolor{gray}{(\textit{arXiv'25})} & 98.2 & 99.8 & 98.2 & 96.1 & 98.1 \\
      % OpenVLA-7B: PIL image path; same selection as the OFT/pi0/pi0.5 blocks (base = the run closest to the
      % three-run mean, AWARE = the best run): base s7 414 / s9 428 / s8 392 / s9 248, AWARE 0.03 s7 432 /
      % 0.02 s7 435 / 0.01 s8 395 / 0.01 s7 286. Averages half-up on raw counts: 1482/2000 = 74.10, 1548/2000 = 77.40.
      OpenVLA~\cite{OpenVLA}~\textcolor{gray}{(\textit{CoRL'24})} & 82.8 & 85.6 & 78.4 & 49.6 & 74.1 \\
      \quad \textbf{\textit{w/} \method}~\textcolor{gray}{(\textit{Ours})} & \textbf{86.4} & \textbf{87.0} & \textbf{79.0} & \textbf{57.2} & \textbf{77.4} \\
      \rowcolor{gray!15} \quad $\Delta$ &
        \textcolor{lightgreen}{\textbf{+3.6}} &
        \textcolor{lightgreen}{\textbf{+1.4}} &
        \textcolor{lightgreen}{\textbf{+0.6}} &
        \textcolor{lightgreen}{\textbf{+7.6}} &
        \textcolor{lightgreen}{\textbf{+3.3}} \\
      \midrule
      $\pi_0$~\cite{pi0}~\textcolor{gray}{(\textit{RSS'25})} & 96.4 & 97.2 & 93.8 & 81.2 & 92.2 \\
      \quad \textbf{\textit{w/} \method}~\textcolor{gray}{(\textit{Ours})} & \textbf{98.2} & \textbf{98.6} & \textbf{94.4} & \textbf{84.0} & \textbf{93.8} \\
      \rowcolor{gray!15} \quad $\Delta$ &
        \textcolor{lightgreen}{\textbf{+1.8}} &
        \textcolor{lightgreen}{\textbf{+1.4}} &
        \textcolor{lightgreen}{\textbf{+0.6}} &
        \textcolor{lightgreen}{\textbf{+2.8}} &
        \textcolor{lightgreen}{\textbf{+1.6}} \\
      \midrule
      $\pi_{0.5}$~\cite{pi0.5}~\textcolor{gray}{(\textit{CoRL'25})} & 98.2 & 98.6 & 97.6 & 92.6 & 96.8 \\
      \quad \textbf{\textit{w/} \method}~\textcolor{gray}{(\textit{Ours})} & \textbf{99.2} & \textbf{99.6} & \textbf{98.4} & \textbf{94.0} & \textbf{97.8} \\
      \rowcolor{gray!15} \quad $\Delta$ &
        \textcolor{lightgreen}{\textbf{+1.0}} &
        \textcolor{lightgreen}{\textbf{+1.0}} &
        \textcolor{lightgreen}{\textbf{+0.8}} &
        \textcolor{lightgreen}{\textbf{+1.4}} &
        \textcolor{lightgreen}{\textbf{+1.0}} \\
      \midrule
      OpenVLA-OFT~\cite{OpenVLA-OFT}~\textcolor{gray}{(\textit{RSS'25})} & 98.0 & 98.2 & 97.6 & 93.8 & 96.9 \\
      \quad \textbf{\textit{w/} \method}~\textcolor{gray}{(\textit{Ours})} & \textbf{99.2} & \textbf{98.6} & \textbf{98.6} & \textbf{96.0} & \textbf{98.1} \\
      \rowcolor{gray!15} \quad $\Delta$ &
        \textcolor{lightgreen}{\textbf{+1.2}} &
        \textcolor{lightgreen}{\textbf{+0.4}} &
        \textcolor{lightgreen}{\textbf{+1.0}} &
        \textcolor{lightgreen}{\textbf{+2.2}} &
        \textcolor{lightgreen}{\textbf{+1.2}} \\
      \bottomrule
    \end{tabular}
  % }
  \label{tab:LIBERO}
\end{table*}

\begin{table*}[t]
  \centering
  \caption{Performance comparison on the SIMPLER benchmark (Visual Matching). Evaluated with official released checkpoints under identical environments and pipelines. Intervention strengths are $\alpha{=}0.09$ for CogACT and $\alpha{=}0.08$ for SpatialVLA. Bold denotes the better of a pair.}
  \label{tab:simpler}
  \newcommand{\hd}[2]{\begin{tabular}[c]{@{}c@{}}\textbf{#1}\\\textbf{#2}\end{tabular}}
  % \resizebox{\linewidth}{!}{%
    \begin{tabular}{lccccc}
      \toprule
      \textbf{Google Robot} & \hd{Pick}{Coke Can} & \hd{Move}{Near} & \hd{Open/Close}{Drawer} & \hd{Open Top Drawer}{and Place Apple} & \textbf{Average} \\
      \midrule
      % Reference rows commented out
      % RT-1~\cite{Rt-1}~\textcolor{gray}{(\textit{arXiv'22})} & 85.7 & 44.2 & 73.0 & 6.5 & 52.4 \\
      % RT-2-X~\cite{OXE}~\textcolor{gray}{(\textit{CoRL'23})} & 78.7 & 77.9 & 25.0 & 3.7 & 46.3 \\
      % OpenVLA~\cite{OpenVLA}~\textcolor{gray}{(\textit{CoRL'24})} & 18.0 & 56.3 & 63.0 & 0.0 & 34.3 \\
      CogACT~\cite{Cogact}~\textcolor{gray}{(\textit{arXiv'24})} & 90.0 & \textbf{85.4} & 70.4 & 49.1 & 73.7 \\
      \quad \textbf{\textit{w/} \method}~\textcolor{gray}{(\textit{Ours})} & \textbf{92.3} & 82.9 & \textbf{75.0} & \textbf{57.4} & \textbf{76.9} \\
      \rowcolor{gray!15} \quad $\Delta$ & \textcolor{lightgreen}{\textbf{+2.3}} & $-$2.5 & \textcolor{lightgreen}{\textbf{+4.6}} & \textcolor{lightgreen}{\textbf{+8.3}} & \textcolor{lightgreen}{\textbf{+3.2}} \\
      \midrule
      \textbf{WidowX Robot} & \hd{Put Spoon}{on Towel} & \hd{Put Carrot}{on Plate} & \hd{Stack Green Block}{on Yellow Block} & \hd{Put Eggplant}{in Yellow Basket} & \textbf{Average} \\
      \midrule
      % Reference rows commented out
      % RT-1-X~\cite{OXE}~\textcolor{gray}{(\textit{CoRL'23})} & 0.0 & 4.2 & 0.0 & 0.0 & 1.1 \\
      % Octo-Base~\cite{Octo}~\textcolor{gray}{(\textit{RSS'24})} & 15.8 & 12.5 & 0.0 & 41.7 & 17.5 \\
      % OpenVLA~\cite{OpenVLA}~\textcolor{gray}{(\textit{CoRL'24})} & 4.2 & 0.0 & 0.0 & 12.5 & 4.2 \\
      SpatialVLA~\cite{SpatialVLA}~\textcolor{gray}{(\textit{RSS'25})} & 16.7 & 25.0 & 29.2 & \textbf{100.0} & 42.7 \\
      \quad \textbf{\textit{w/} \method}~\textcolor{gray}{(\textit{Ours})} & \textbf{20.8} & \textbf{29.2} & \textbf{33.3} & 95.8 & \textbf{44.8} \\
      \rowcolor{gray!15} \quad $\Delta$ & \textcolor{lightgreen}{\textbf{+4.2}} & \textcolor{lightgreen}{\textbf{+4.2}} & \textcolor{lightgreen}{\textbf{+4.2}} & $-$4.2 & \textcolor{lightgreen}{\textbf{+2.1}} \\
      \bottomrule
    \end{tabular}
\end{table*}

\subsection{Simulation Experiments}
\label{sec:simulation}

\noindent\textbf{Simulation Setup and Baselines.}
To systematically validate \method across diverse robotic manipulation settings, we evaluate on three widely-adopted simulation benchmarks: LIBERO~\cite{libero}, SIMPLER~\cite{simpler}, and CALVIN~\cite{calvin}. Across these environments, we evaluate seven representative VLA backbones spanning model scales from 0.5B to 7B parameters: single-system models, including LLaVA-VLA (0.5B)~\cite{LLaVA-VLA}, SpatialVLA (4B)~\cite{SpatialVLA}, OpenVLA (7B)~\cite{OpenVLA}, and OpenVLA-OFT (7B)~\cite{OpenVLA-OFT}; and dual-system architectures, including $\pi_0$ (3B)~\cite{pi0}, $\pi_{0.5}$ (3B)~\cite{pi0.5}, and CogACT (7B)~\cite{Cogact}. All base policies and \method are evaluated on NVIDIA A100 GPUs using official checkpoints under identical environments, initial states, and pipelines. In all experiments, \method operates entirely training-free at test time without additional observation cues or learnable parameters. To reflect practical deployment across tasks with diverse horizons and precision demands, the intervention strength $\alpha \in [0.01, 0.10]$ is calibrated per task suite, with ablation sweeps in Section~\ref{sec:ablation} confirming that policy gains remain robust across a broad interior range.

\noindent\textbf{LIBERO.} Table~\ref{tab:LIBERO} reports policy performance across the four suites of the LIBERO benchmark~\cite{libero} (Spatial, Object, Goal, and Long), each comprising ten manipulation tasks tested over 500 episodes per suite. Across all four heterogeneous VLA backbones, \method consistently improves manipulation success over untouched base policies. Because extended temporal horizons exacerbate visual representation fading, the benefits of \method are especially pronounced on LIBERO-Long: boosting OpenVLA by an absolute 7.6\%, $\pi_0$ by 2.8\%, $\pi_{0.5}$ by 1.4\%, and OpenVLA-OFT to 96.0\%. On near-saturated suites such as Spatial, Object, and Goal, the uncertainty-guided weighting naturally scales down intervention on confident tokens, steadily advancing performance toward ceiling (e.g., reaching 99.2\% on Spatial) without inducing regressions. These across-the-board gains confirm that associative observation retrieval provides a robust, training-free mechanism benefiting both autoregressive and flow-matching policies.

\noindent\textbf{SIMPLER.} As summarized in Table~\ref{tab:simpler}, we assess cross-embodiment generalization under visual matching on the SIMPLER benchmark~\cite{simpler} across two distinct robot platforms: dual-system CogACT (7B)~\cite{Cogact} on Google Robot and SpatialVLA (4B)~\cite{SpatialVLA} on WidowX BridgeData. On Google Robot, \method raises CogACT's average success to 76.9\%, with the most substantial boost (+8.3\%) achieved on the multi-stage task \textit{Open Top Drawer and Place Apple}, where precise visual relocalization is essential for gripper alignment. On WidowX, \method lifts SpatialVLA to 44.8\% across 96 paired evaluations, achieving seven wins against five losses across initial configurations. These gains confirm that \method generalises seamlessly across diverse robot kinematics, observation viewpoints, and control frequencies without modifying official deployment pipelines.

\begin{table}[t]
  \centering
  \setlength{\tabcolsep}{1mm}
  \caption{Performance comparison on the CALVIN ABC$\rightarrow$D benchmark, in success rate per chain length and average completed length. Evaluated with the released checkpoint; a single strength $\alpha{=}0.01$ is used across the entire benchmark.}
  \label{tab:CALVIN}
  \resizebox{\linewidth}{!}{%
    \begin{tabular}{lcccccc}
      \toprule
      \multirow{2}{*}{\textbf{Method}} & \multicolumn{5}{c}{\textbf{Success Rate (\%)}} & \multirow{2}{*}{\textbf{Avg. Len}} \\
      & \textbf{1/5} & \textbf{2/5} & \textbf{3/5} & \textbf{4/5} & \textbf{5/5} & \\
      \midrule
      % Reference baselines commented out
      % RoboFlamingo~\cite{li2024roboflamingo}~\textcolor{gray}{(\textit{ICLR'24})} & 82.4 & 61.9 & 46.6 & 33.1 & 23.5 & 2.47 \\
      % VidMan~\cite{vidman}~\textcolor{gray}{(\textit{NeurIPS'24})} & 91.5 & 76.4 & 68.2 & 59.2 & 46.7 & 3.42 \\
      % OpenVLA~\cite{OpenVLA}~\textcolor{gray}{(\textit{CoRL'24})} & 91.3 & 77.8 & 62.0 & 52.1 & 43.5 & 3.27 \\
      % VLAS~\cite{vlas}~\textcolor{gray}{(\textit{ICLR'25})} & 87.2 & 64.2 & 40.9 & 28.1 & 19.6 & 2.40 \\
      LLaVA-VLA~\cite{LLaVA-VLA}~\textcolor{gray}{(\textit{ICRA'26})} & 94.0 & 82.4 & 68.2 & 57.8 & 47.0 & 3.49 \\
      \quad \textbf{\textit{w/} \method}~\textcolor{gray}{(\textit{Ours})} & \textbf{94.4} & \textbf{82.8} & \textbf{72.4} & \textbf{62.8} & \textbf{50.4} & \textbf{3.63} \\
      \rowcolor{gray!15} \quad $\Delta$ &
        \textcolor{lightgreen}{\textbf{+0.4}} &
        \textcolor{lightgreen}{\textbf{+0.4}} &
        \textcolor{lightgreen}{\textbf{+4.2}} &
        \textcolor{lightgreen}{\textbf{+5.0}} &
        \textcolor{lightgreen}{\textbf{+3.4}} &
        \textcolor{lightgreen}{\textbf{+0.14}} \\
      \bottomrule
    \end{tabular}}
\end{table}

\begin{figure*}[t]
  \centering
  \includegraphics[width=\linewidth]{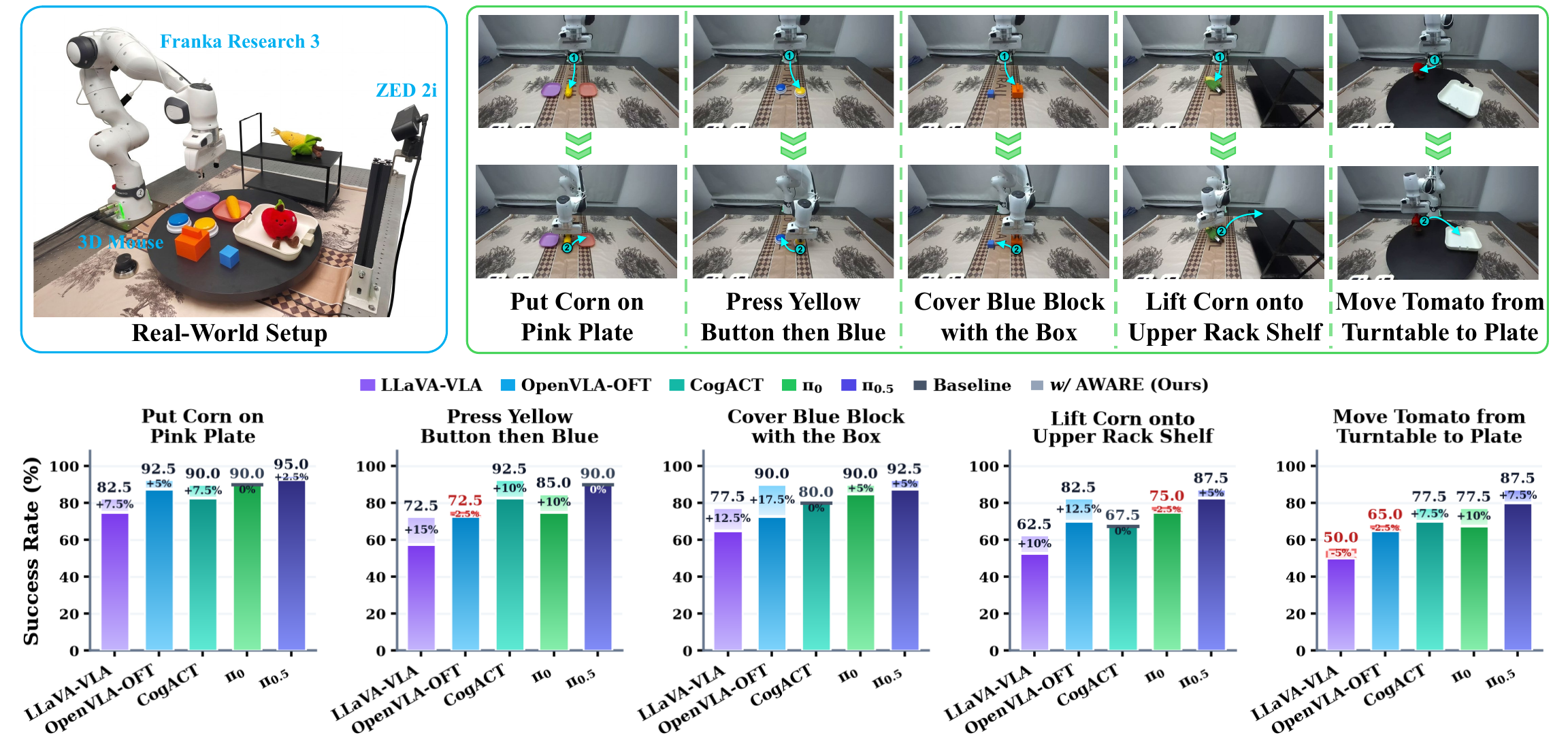}
  \caption{Real-world experimental setup and manipulation results. Top: Franka Research 3 robot workstation (left) and the five evaluated manipulation tasks (right). Bottom: Task success rates across five representative VLA backbones comparing baseline policies with \textit{w/} \method over 40 physical rollouts per task (2,000 trials total). Intervention strengths are $\alpha{=}0.01$ for LLaVA-VLA, $\alpha{=}0.04$ for OpenVLA-OFT, $\alpha{=}0.05$ for CogACT, and $\alpha{=}0.03$ for $\pi_0$ and $\pi_{0.5}$.}
  \label{fig:real_robot}
\end{figure*}

\label{sec:calvin}
\noindent\textbf{CALVIN.} Table~\ref{tab:CALVIN} evaluates multi-stage sequential manipulation on the CALVIN ABC$\rightarrow$D benchmark~\cite{calvin}, testing zero-shot generalization across 500 language-conditioned task chains with the compact LLaVA-VLA backbone~\cite{LLaVA-VLA}. Although base policies achieve high initial success on early subtasks, compounding spatial drift typically causes steep failure cascades at deeper chain stages. Crucially, \method provides increasingly pronounced gains as task depth increases, mitigating compounding failure cascades and boosting subtask success by 4.2\% at chain length 3, 5.0\% at length 4, and 3.4\% at length 5. Consequently, the complete five-subtask chain completion rate rises to 50.4\%, and the average completed chain length increases to 3.628 subtasks. This confirms that targeted associative retrieval effectively counteracts perceptual fading over extended horizons, even within a compact 0.5B parameter policy.

\subsection{Real-World Experiments}
\label{sec:real-world}

\noindent\textbf{Real-World Setup.}
We perform real-robot experiments to validate the effectiveness of \method in the real world. As illustrated in Fig.~\ref{fig:real_robot} (top), our workstation comprises a Franka Research 3 manipulator with a parallel-jaw gripper, a static overhead ZED~2i camera, and a 3Dconnexion 3D mouse. We evaluate on five diverse tasks: 1)~\textit{Put Corn on Pink Plate}, 2)~\textit{Press Yellow Button then Blue}, 3)~\textit{Cover Blue Block with the Box}, 4)~\textit{Lift Corn onto Upper Rack Shelf}, and 5)~\textit{Move Tomato from Turntable to Plate}, spanning short-horizon placement, sequential button pressing, obstacle occlusion, multi-tier precision lifting, and dynamic moving-target acquisition. For each task, we collect 50 teleoperated demonstrations to fine-tune in-domain checkpoints. We evaluate five representative policy backbones (LLaVA-VLA, OpenVLA-OFT, CogACT, $\pi_0$, and $\pi_{0.5}$) across 40 physical rollouts per task under randomized layouts and initial poses, totaling 2,000 real-world evaluation trials.

\noindent\textbf{Results.}
Figure~\ref{fig:real_robot} summarizes physical evaluation results across all five policy backbones under a fixed intervention strength per model family. Across all architectures, \method consistently enhances physical manipulation, yielding average success gains ranging from +4.0\% on high-capacity $\pi_{0.5}$ to +8.0\% on compact LLaVA-VLA. Out of 25 evaluated task-model combinations, \method achieves positive improvements in 17 settings, reaching an absolute peak of 95.0\% on \textit{Put Corn on Pink Plate} with $\pi_{0.5}$ and a maximum single-task boost of +17.5\% on \textit{Cover Blue Block with the Box} using OpenVLA-OFT. Lower-capacity policies gain substantial headroom on challenging contact-rich tasks (e.g., a 15.0\% improvement on sequential button pressing with LLaVA-VLA), while saturated policies maintain steady progress near performance ceilings. Neutral outcomes in four boundary cases and slight trade-offs under high-speed turntable dynamics reflect authentic physical challenges under fixed parameters, confirming that \method delivers generalizable, training-free robustness in real-world robotic manipulation.

\subsection{Ablation Studies}
\label{sec:ablation}
\begin{figure}[t]
  \centering
  \begin{subfigure}[b]{0.49\linewidth}
    \centering
    \includegraphics[width=\linewidth]{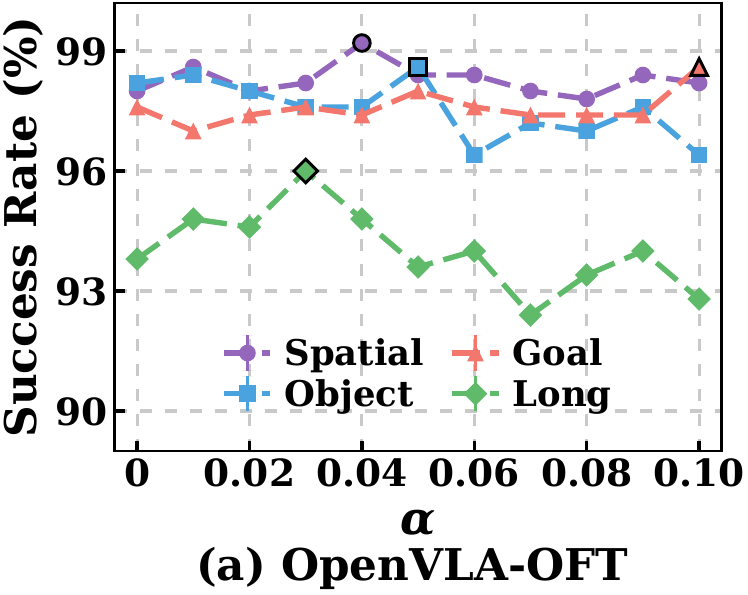}
    \label{fig:alpha_oft}
  \end{subfigure}\hfill
  \begin{subfigure}[b]{0.49\linewidth}
    \centering
    \includegraphics[width=\linewidth]{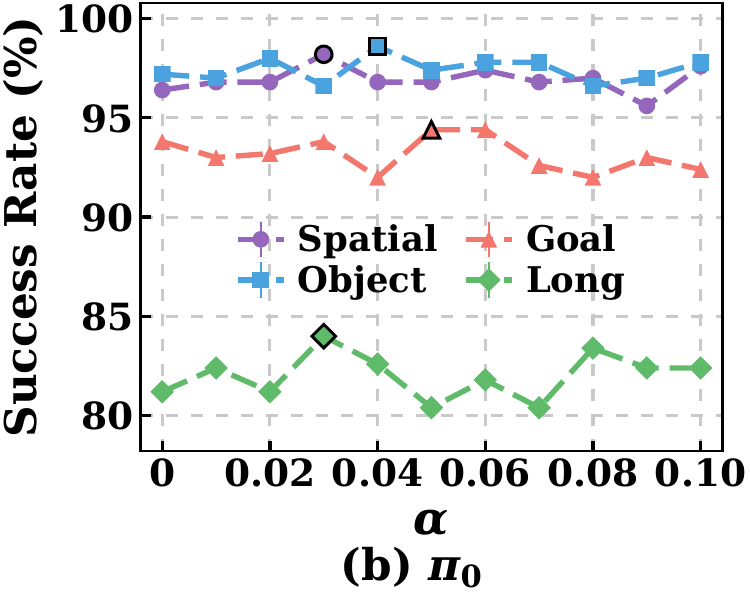}
    \label{fig:alpha_pi0}
  \end{subfigure}
  \caption{\textbf{Impact of the strength $\alpha$.} Success rate per suite across the parameter grid; $\alpha{=}0$ denotes the untouched baseline policy, and outlined markers denote the optimal operating points for each suite (reported in Table~\ref{tab:LIBERO}).}
  \label{fig:alpha}
\end{figure}

Focusing on OpenVLA-OFT, our ablation studies investigate three core aspects of \method: component contributions in Table~\ref{tab:ablation}, mechanistic validations in Table~\ref{tab:controls}, and sensitivity to the intervention strength $\alpha$ in Fig.~\ref{fig:alpha}.

\begin{table}[t]
\centering
% \small
\setlength{\tabcolsep}{3.4pt}
\caption{Component analysis on the LIBERO benchmark with OpenVLA-OFT using the baseline strengths from Table~\ref{tab:LIBERO}. Ablating individual components isolates the contribution of each design choice while holding the global intervention scale $\alpha$ fixed.}
\label{tab:ablation}
\resizebox{\linewidth}{!}{%
\begin{tabular}{ccc@{\hskip 4mm}ccccc}
\toprule
\multicolumn{3}{c}{\textbf{Part of the rule kept}} & \multicolumn{5}{c}{\textbf{Success rate (\%)}} \\
\cmidrule(r){1-3} \cmidrule(l){4-8}
\makecell{\textbf{Per-element}\\\textbf{weight}} & \makecell{\textbf{Depth}\\\textbf{factor}} & \makecell{\textbf{Instruction}\\\textbf{guidance}} & \textbf{Spatial} & \textbf{Object} & \textbf{Goal} & \textbf{Long} & \textbf{Average} \\
\midrule
$\times$ & $\times$ & $\times$ & 98.0 & 98.2 & 97.6 & 93.8 & 96.9 \\
\midrule
$\times$ & \checkmark & \checkmark & 98.8 & 97.0 & 98.4 & 95.2 & 97.4 \\
\checkmark & $\times$ & \checkmark & 98.2 & 97.2 & 97.8 & 94.8 & 97.0 \\
\checkmark & \checkmark & $\times$ & 98.6 & 96.6 & 98.4 & 95.2 & 97.2 \\
\rowcolor{gray!15} \checkmark & \checkmark & \checkmark & \textbf{99.2} & \textbf{98.6} & \textbf{98.6} & \textbf{96.0} & \textbf{98.1} \\
\bottomrule
\end{tabular}}
\end{table}

\begin{table}[t]
\centering
\setlength{\tabcolsep}{1.5mm}
\caption{Diagnostic variants on OpenVLA-OFT (LIBERO-Long) testing alternative explanations across 500 evaluation episodes.}
\label{tab:controls}
\resizebox{\linewidth}{!}{%
\begin{tabular}{llcc}
\toprule
  \textbf{Tested aspect} & \textbf{Evaluated variant} & \textbf{Succ. (\%)} & $\boldsymbol{\Delta}$ \textbf{(\%)} \\
\midrule
\multirow{2}{*}{Reference} & OpenVLA-OFT (Base) & 93.8 & -- \\
 & \cellcolor{gray!15}\textbf{\method (Ours)} & \cellcolor{gray!15}\textbf{96.0} & \cellcolor{gray!15}\textcolor{lightgreen}{\textbf{+2.2}} \\
\midrule
\multirow{2}{*}{Uncertainty routing} & Linear chunk schedule ($w \propto k/H$) & 94.0 & +0.2 \\
 & Permuted entropy weights & 94.8 & +1.0 \\
\midrule
\multirow{3}{*}{Injected content} & Unrelated episode memory & 94.8 & +1.0 \\
 & Matched Gaussian noise & 94.6 & +0.8 \\
 & FFN damping w/o injection & 94.0 & +0.2 \\
\bottomrule
\end{tabular}}
\end{table}

\noindent\textbf{Component Analysis.} Table~\ref{tab:ablation} isolates each core component of \method with global scale $\alpha$ fixed. The full method achieves 98.1\% average success, with all elements contributing complementarily. Replacing element-specific entropy with uniform weights reduces success to 97.4\%, confirming the necessity of fine-grained allocation. Ablating the depth schedule $\ell/(L{-}1)$ causes the sharpest drop (to 97.0\%), demonstrating that intervention is crucial where observation grounding decays. Omitting instruction conditioning reduces success to 97.2\%, verifying its distractor filtering role. Across suites, LIBERO-Object is most sensitive (dropping $\ge 1.4\%$ under any ablation), reflecting strict reliance on precise perceptual grounding.

\noindent\textbf{Mechanistic Analysis.} To isolate the driver of policy gains, Table~\ref{tab:controls} benchmarks five diagnostic variants on LIBERO-Long. First, replacing Action Entropy with a linear chunk schedule ($w \propto k/H$) yields only +0.2\%, showing dynamic uncertainty cannot be replaced by a fixed schedule. Second, permuting entropy weights across action dimensions halves the gain (+1.0\%), confirming features must target genuinely uncertain dimensions. Third, retrieving memory features from an unrelated episode recovers only +1.0\%, showing generic visual tokens do not suffice. Fourth, injecting norm-matched Gaussian noise yields merely +0.8\%, ruling out random perturbation. Finally, retaining native FFN damping ($(1{-}w)\operatorname{FFN}$) without injection gives +0.2\%, showing suppression alone is insufficient. Together, these results confirm that \method specifically routes authentic visual evidence to high-entropy action dimensions.

\noindent\textbf{Sensitivity to Strength $\alpha$.} Sweeping $\alpha \in [0.01, 0.10]$ for OpenVLA-OFT and $\pi_0$ (Fig.~\ref{fig:alpha}) reveals a shared optimum at $\alpha{=}0.03$ on LIBERO-Long (+2.2\% and +2.8\%, respectively). Crucially, adopting a single default $\alpha{=}0.03$ across suites outperforms base policies without per-task tuning, confirming gains do not rely on oracle search. Saturated suites remain stable ($\alpha \in [0.01, 0.05]$), while larger strengths ($\alpha \ge 0.06$) degrade precision. Autoregressive models similarly favor conservative scales ($\alpha \le 0.03$ for OpenVLA, $\alpha{=}0.01$ for LLaVA-VLA) to avoid over-perturbation.

\subsection{Inference Overhead}
\label{sec:cost}
\begin{table}[t]
  \centering
  \caption{Inference latency and throughput on OpenVLA-OFT (LIBERO-Long) evaluated on a single NVIDIA A100 GPU. Because the policy predicts action chunks of horizon $H=8$, per-step latency reflects the effective wait time per environment interaction step.}
  \label{tab:overhead}
  % \resizebox{\linewidth}{!}{%
\begin{tabular}{lccc}
    \toprule
    \textbf{Method} & \textbf{Throughput} $\uparrow$ & \textbf{Per call} $\downarrow$ & \textbf{Per env.\ step} $\downarrow$ \\
    \midrule
    OpenVLA-OFT & 82.9 Hz & 96.5 ms & 29.4 ms \\
    \quad w/ \method & 68.1 Hz & 117.5 ms & 32.0 ms \\
    % \midrule
    \quad $\Delta$ & $-17.9\%$ & $+21.8\%$ & $+8.9\%$ \\
    \bottomrule
  \end{tabular}
\end{table}

\method introduces minimal computational overhead by design. Instruction weighting $\boldsymbol{g}_t$ is computed once per policy call, and observation features $\boldsymbol{O}_t$ are pre-cached. At each layer, computing $\mathcal{H}^{(\ell)}_{t,n}$ and associative retrieval operate strictly on the compact action subset ($N_a \ll N$, e.g., 56 action elements vs.\ $\sim$1000 multimodal tokens). Rather than evaluating heavy self-attention across all $N$ tokens, \method requires only two lightweight matrix multiplications per layer. Benchmarks in Table~\ref{tab:overhead} confirm that on OpenVLA-OFT, \method adds merely 2.6\,ms per environment interaction step ($+8.9\%$, 29.4\,ms to 32.0\,ms), sustaining 68.1\,Hz throughput and preserving real-time control feasibility.

\section{Conclusion}
\label{sec:conclusions}
We have presented \method, a lightweight, training-free associative memory retrieval framework for Vision-Language-Action (VLA) models. Motivated by the observation that action representations lose observation grounding across depth while predictive uncertainty converges asynchronously across dimensions, \method employs an element-wise Action Entropy metric to selectively retrieve instruction-guided observation evidence into high-entropy action dimensions via native feed-forward networks. Across seven diverse VLA backbones on LIBERO, SIMPLER, CALVIN, and real-world robots, \method consistently improves task execution success on extended-horizon manipulation with negligible computational overhead, offering a practical plug-and-play foundation for robust robot control.

% \newpage
{\footnotesize
\bibliographystyle{IEEEtranN}
\bibliography{references}
}
\end{document}